\documentclass{iet-ell}

\usepackage{algorithmic}
\usepackage{graphicx}
\usepackage{diagbox}
\usepackage{subcaption}

\usepackage[none]{hyphenat}
\usepackage{tikz}
\usepackage[absolute,overlay]{textpos}
\usepackage{blindtext}
\usepackage{relsize}
\usepackage{hyperref}

\tikzset{fontscale/.style={font=\relsize{#1}}}

\def\BibTeX{{\rm B\kern-.05em{\sc i\kern-.025em b}\kern-.08em
    T\kern-.1667em\lower.7ex\hbox{E}\kern-.125emX}}

\setcopyright{open}
\ietVolume{61}
\ietYear{2025}
\ietDoi{ell2.70210}

\received{2 January 2025}
\accepted{8 March 2025}

\begin{document}

\title{Balancing Robustness and Efficiency in Embedded DNNs Through Activation Function Selection}

\author[af1]{Jon Gutiérrez-Zaballa}
\orcid{0000-0002-6633-4148}
\author[af1]{Koldo Basterretxea}
\orcid{0000-0002-5934-4735}
\author[af2]{Javier Echanobe}
\orcid{0000-0002-1064-2555}

\affil[af1]{Department of Electronics Technology, University of the Basque Country (UPV/EHU), 48013, Bilbao, Spain}
\affil[af2]{Department of Electricity and Electronics, University of the Basque Country (UPV/EHU), 48940, Leioa, Spain}

\corresp{Email: j.gutierrez@ehu.eus}

\begin{abstract}%
Machine learning-based embedded systems for safety-critical applications, such as aerospace and autonomous driving, must be robust to perturbations caused by soft errors.
As transistor geometries shrink and voltages decrease, modern electronic devices become more susceptible to background radiation, increasing the concern about failures produced by soft errors.
The resilience of deep neural networks (DNNs) to these errors depends not only on target device technology but also on model structure and the numerical representation and arithmetic precision of their parameters.
Compression techniques like pruning and quantization, used to reduce memory footprint and computational complexity, alter both model structure and representation, affecting soft error robustness.
In this regard, although often overlooked, the choice of activation functions (AFs) impacts not only accuracy and trainability but also compressibility and error resilience.
This paper explores the use of bounded AFs to enhance robustness against parameter perturbations, while evaluating their effects on model accuracy, compressibility, and computational load with a technology-agnostic approach.
We focus on encoder-decoder convolutional models developed for semantic segmentation of hyperspectral images with application to autonomous driving systems.
Experiments are conducted on an AMD-Xilinx’s KV260 SoM.
\end{abstract}

\maketitle
\begin{textblock*}{21cm}(1.5cm,28.5cm)
  \begin{tikzpicture}
    \draw (0,0) rectangle (17.75,0.5); 
    \end{tikzpicture}
\end{textblock*} 

\begin{textblock*}{21cm}(0cm,28.5cm)
  \begin{tikzpicture}
    \node (center) {c};
    \path (center)+(10.5,4) node [fontscale=-1] (name) {\copyright 2025 Wiley. Final published version of the article can be found at \href{https://ietresearch.onlinelibrary.wiley.com/doi/10.1049/ell2.70210}{10.1049/ell2.70210}.};
    \end{tikzpicture}
\end{textblock*}

\section{Introduction}\label{sec:IntroRelatedWork}
Deploying AI accelerators on the edge enhances security, reliability, and latency by eliminating external data communication.
This provides full execution autonomy for complex deep neural networks (DNNs), making them suitable for real-time, safety-critical applications with strict communication and reliability standards, such as autonomous driving systems (ISO 26262) and aerospace (ARP 4754).
A key reliability concern is the exposure of electronic components to background radiation, especially in memory devices with small charge storage and large silicon areas.
Even a minor perturbation in a memory cell can cause a single event upset (SEU).

Field-programmable gate arrays (FPGAs), widely used in these domains, are particularly vulnerable to SEUs, as both configuration memory (LUTs) and sequential elements (flip-flops or Block RAMs) can be affected.
Without proper protection, the mean time between failures could drop to just a few seconds \cite{deviceReliabilityReport}.
Furthermore, DNNs often depend on highly complex models with millions of parameters and tens of giga floating-point operations (GFLOPS) per inference.
Deploying them on resource-constrained embedded processors requires compression techniques to balance accuracy, hardware usage, and inference speed.

In this paper, we extend the previous work in \cite{10769119} to include a more detailed numerical analysis on the suitability of using bounded activation functions (AFs) as a means to improve the robustness of image segmentation DNNs in the context of autonomous driving systems.
We include a brief discussion section where we analyse the impact of the AF choice on the performance and FPGA implementability of these models.

The article is structured as follows: First, we review SotA studies on DNN vulnerabilities to soft errors and hardening strategies.
Next, we outline DNN development, focusing on AF selection and compression techniques.
We then analyse the fault injection (FI) campaign to assess the impact of compression on different AF-based DNN robustness.
After that, the FPGA deployment process is characterized.
Finally, we present conclusions relevant to similar AI deployments.

\section{State of the art}\label{sec:StateOfTheArt}
The vulnerabilities of DNNs have been studied mainly through simulated software FI campaigns.
However, some researches have used radiation tests or theoretical analyses based on error propagation models \cite{systematicReview}.
Nevertheless, most studies focus on image classification rather than segmentation, highlighting the need for further research to guide the efficient design of segmentation models.

FI campaign analyses require a statistically significant approach with numerous per-layer FIs.
Given the computational complexity of SotA CNNs, \cite{ruospo2023assessing} proposes a data-aware optimization to reduce FIs while maintaining accuracy.
One of the most extensive image-classification CNN analyses \cite{hong2019terminal} evaluates a 32-bit floating-point CNN's vulnerability based on bit position, flip direction, and sign.
Studies \cite{ruospo2021investigating, syed2021fault} show that fixed-point representation enhances resilience and reduces model complexity, with heterogeneous quantization further improving robustness.
Lastly, \cite{Dong_2023_ICCV} examines adversarial attacks, identifying a bit-flip threshold that renders models useless.
Few studies focus on segmentation DNNs.
\cite{govarini2023fast} highlights the need for statistical significance by comparing two unstructured FI campaigns with a properly defined one on U-Net.
\cite{burel2024techniques} identifies critical single bit-flips affecting floating-point DeepLabV3+ robustness.
\cite{iurada:hal-04684784} also employs DeepLabV3 with a convolution fault model derived from beam experiments and RTL fault injections.
\cite{GUTIERREZZABALLA2024103242} conducts a layer-by-layer, bit-by-bit vulnerability analysis of U-Net, examining the impact of compression techniques like pruning and quantization on robustness.

Researchers are developing methods to harden DNNs against soft errors, categorized into redundancy, parameter modification, AF modification, and training modification.
While redundancy techniques like full duplication and triple modular redundancy are effective, their high memory overhead limits embedded applications.
To address this, optimized redundancy methods targeting critical parameters have been proposed in \cite{ruospo2022selective, bolchini2022selective}.
Parameter modification methods aim for zero memory overhead.
MATE \cite{jang2021mate} is an error correction tool that leverages the least significant bits of weight mantissas for error correction.
More recently, \cite{GUTIERREZZABALLA2024103242} introduced an overhead-free approach that selectively protects model parameters, addressing challenges like partial exponent completion.
AF modification provides a low-overhead hardening approach.
\cite{hoang2020ft} proposed zeroing ReLU outputs when inputs exceed a threshold to mitigate faulty weight effects.
To reduce FI campaign costs, \cite{burel2024techniques} estimated thresholds using activation statistics (\textit{min}, \textit{avg}, \textit{max}, \textit{std}).
Instead of zeroing activations, \cite{taheri2024exploration} explored replacing AF maximums with bound values.
However, the impact of AF choices on performance and on-the-edge implementability remains underexplored.
Finally, articles such as \cite{9897813, gambardella2022accelerated} have explored fault-aware training with positive results.
This implies strategically corrupting some feature maps during training so that the DNN learns how to tackle them.

\section{Reference DNN model}\label{sec:modelDevelopment}
The reference DNN in this study is a U-Net adapted to be trained with hyperspectral images from the HSI-Drive v2.0 dataset \cite{gutierrez2023hsi}.
The resulting model features 31.14 Mparameters (249.04 MB) and requires 34.60 GFLOPS per inference (see \cite{10769119}).

AFs with unbounded output ranges, such as ReLU, are prone to propagating errors from SBUs, as the impact of bit flips cannot be controlled.
This may result in extreme shifts in the DNN model’s behaviour.
This study aims to investigate how effectively bounded AFs can mitigate this issue.
To analyse both model robustness and implementation performance, we selected the Sigmoid (Sig) AF and its more computationally efficient variant, Hard Sigmoid (HSig), depicted in Equation \ref{equ:HardSig}.
Both AFs limit the output range to [0, 1], helping control the influence of perturbations.

\begin{equation}
HSig(x) =
\begin{cases}
0 & \text{if } x \leq -3 \\
1 & \text{if } x \geq 3 \\
\frac{x}{6} + 0.5 & \text{if } -3 < x < 3
\end{cases}
\label{equ:HardSig}
\end{equation}

The evaluation of the DNNs on the test set has produced the following results (Global IoU, GIoU / Weighted IoU): (94.71/88.54) for the ReLU, (93.32/84.75) for the Sig, and (92.02/82.69) for the HSig.
All metrics are above (92/82) and are considered satisfactory for the dataset.
As for the training process, the simplicity of the ReLU AF has resulted in a lower number of required epochs (200) compared to the Sig and HSig (1000).

\subsection{Model compression}\label{sec:pruning}
Training large, sparse DNNs and compressing them is the prevalent approach to optimize AI models, as it typically yields better performance than training smaller, denser models.

In the two-step iterative filter pruning process, the pruning criterion is first selected, primarily aiming to reduce FLOPs.
A sensitivity analysis then identifies the most suitable filters for pruning by gradually pruning each layer (0 to 0.9 in steps of 0.1) while freezing others.
A parameter's sensitivity is assessed using the L1 norm, prioritizing less sensitive filters.
The last layer is excluded to maintain a constant number of segmented classes.
Pruning ratios for each iteration are determined through successive sensitivity analyses, leading to an overall pruning ratio with layer-specific adjustments.
After pruning and fine-tuning, the DNN is evaluated against a stopping criteria ($\leq$ 1.5-point GIoU/WIoU degradation).
If unmet, additional pruning iterations follow.
After two iterations, computational complexity (pruning ratio/Mparams/GFLOPS) is: ReLU (0.75/0.32/8.41), Sig (0.52/1.38/16.46) and HSig (0.7/1.47/10.35).
The Sig-based DNN has fewer parameters than the HSig-based DNN due to more aggressive pruning near the base, where most parameters reside (see \cite{10769119}).

After pruning, quantization was applied to reduce the memory usage and accelerate on-the-edge DNN inference.
As the robustness analysis is based on simulations and AMD-Xilinx's Deep Processing Unit (DPU) cannot be modified to emulate faults in its processing elements, we used \textit{TensorFlow Lite}'s general quantization scheme: 8-bit and 32-bit integer representations for weights and biases respectively.
This analysis is relevant for both FPGA and embedded GPU systems, where similar quantization schemes may be applied or quantization might not be needed.
For FPGA implementation, a homogeneous post-training quantization converted all parameters and activations to 8-bit integers.
To ensure fairness, we verified that the observed variation in GIoU between both schemes was below $\pm 0.08$.
After quantization (details available in \cite{10382745}), the Sig-based DNN suffered significant accuracy loss, unlike the ReLU- and HSig-based models, which benefited from Cross Layer Equalization.
Since this technique requires a piecewise linear AF, Sig does not meet this condition and a fast fine-tuning process was performed to recover the lost accuracy.

\section{Analysis of robustness against SBUs}\label{sec:robustness}
TensorFI2 \cite{tensorfi2}, modified to support integer-quantized DNNs (see \textit{\url{https://github.com/jonGuti13/TensorFI2}}), has been used as the software injection tool.
To ensure statistically significant results, the number of injected single-bit flips per-layer has been adjusted according to Equation \ref{equ:statisticallySignificant} from \cite{statisticalFaultInjection}

\vspace{-0.1cm}
\begin{equation}
    n = \frac{N}{1 + e^2 * \frac{N-1}{t^2 * p * (1-p)}}
    \label{equ:statisticallySignificant}
\end{equation}

\begin{figure}[b]
\centerline{\includegraphics[height =6.4cm]{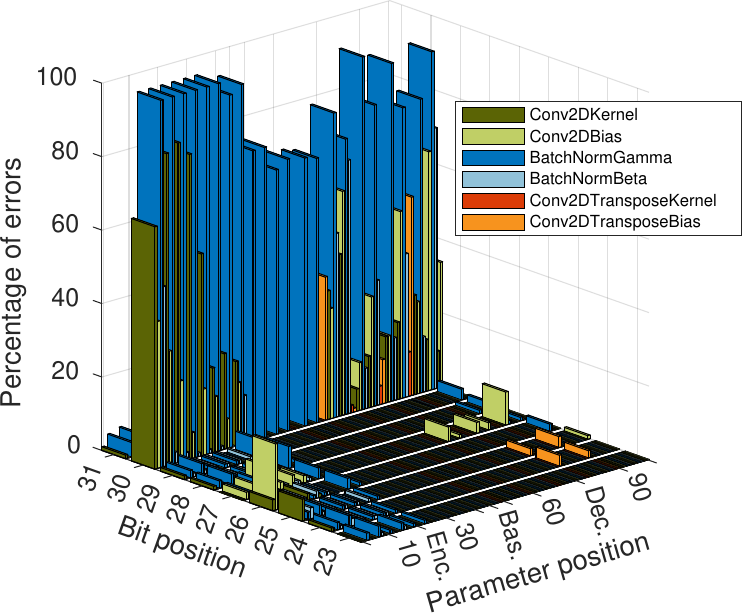}}
\caption{Bit-flip error rate for the original ReLU-based DNN.}
\label{fig:bitFlipErrorReLUNotPruned}
\end{figure}

In Equation \ref{equ:statisticallySignificant}, $N$ represents the number of possible faults per layer (calculated as the number of parameters multiplied by the parameter bit-width), $e$ (set to 0.025) is the 2.5\% error margin, $t$ (set to 1.96) is the value for a 95\% confidence level, and $p$ (set to 0.5) is the probability of faults causing failure.
Since $p$ is unknown, the most conservative approach is to maximize the sample size and assume a 50\% probability, as done in \cite{gambardella2022accelerated}.
$n$ denotes the minimum number of injections for statistical significance, capped at 1550 for the chosen parameters, as in our previous studies \cite{10769119, GUTIERREZZABALLA2024103242}.
For the error metrics, a failure is defined as when any pixel’s predicted class in the output image differs from the unperturbed golden reference.
A 100\% rate represents the situation were all pixels in an image have changed.

\begin{figure}[t]
\centerline{\includegraphics[height =6.3cm]{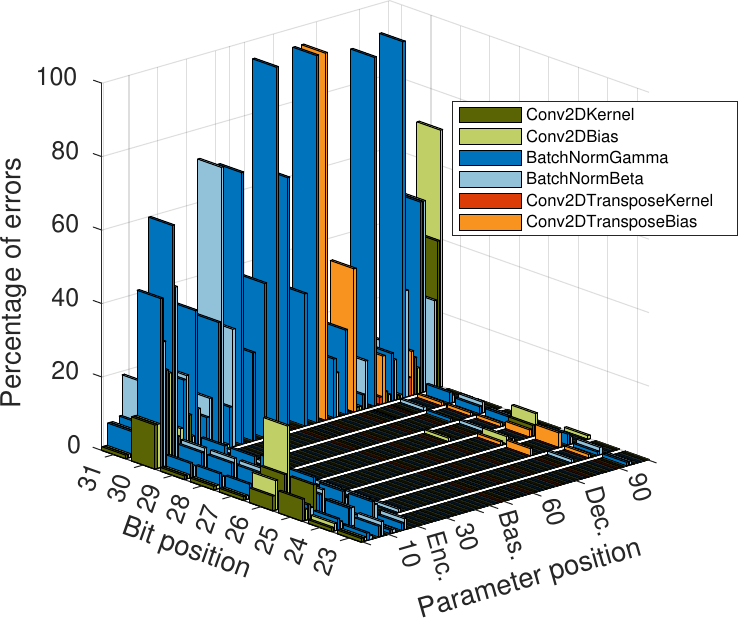}}
\caption{Bit-flip error rate for the original Sig.-based DNN.}
\label{fig:bitFlipErrorSigmoidNotPruned}
\end{figure}

\begin{figure}[b]
\centerline{\includegraphics[height =6.3cm]{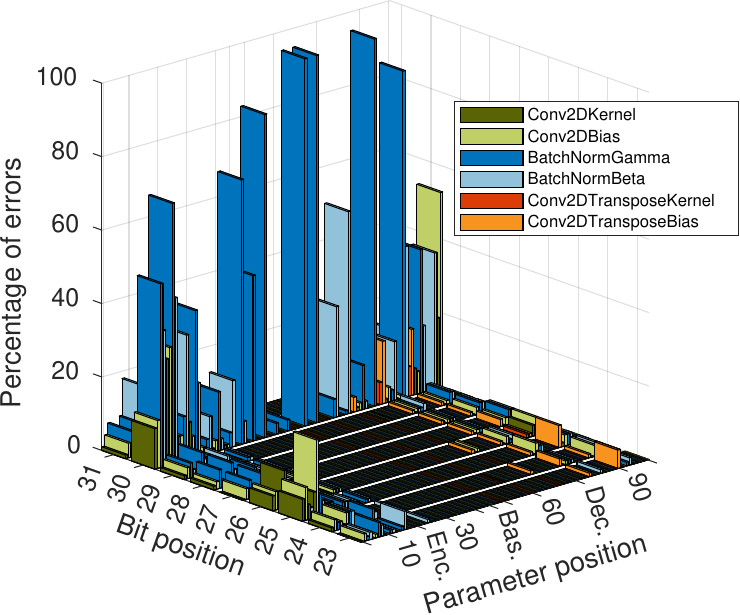}}
\caption{Bit-flip error rate for the original HSig-based DNN.}
\label{fig:bitFlipErrorHardSigmoidNotPruned}
\end{figure}

\subsection{Robustness of noncompressed models}
Figures \ref{fig:bitFlipErrorReLUNotPruned}, \ref{fig:bitFlipErrorSigmoidNotPruned} and \ref{fig:bitFlipErrorHardSigmoidNotPruned}  display the error rate for each layer and for each flipped bit position for the three noncompressed 32-bit floating-point DNN models under study.
As mantissa bits, that is, bit positions below 23, have generated a negligible amount of errors, they have been omitted.

A common error pattern can be observed there, which is a consequence of the data representation and the DNN architecture.
The most destructive situation occurs when the MSB of the exponent (bit 30) flips from `0' to `1', causing a large increase in the represented value.
This is common in these models, as over 99.9\% of parameters fall within the $0 \leq |x| < 2$ range.
A particularly risky situation occurs when the parameter value is in the $1 \leq |x| < 2$ range, as the flip can result in $\pm \infty$ or a $NaN$.
A `0' to `1' flip in other exponent bits can also result in non-negligible error rates, as seen with bit 26 of layer 2 in Figure \ref{fig:bitFlipErrorReLUNotPruned}.
This can be explained because 97\% of the parameters of that layer have a `0' in that position.
This effect can be accentuated when the partial exponent of a parameter \cite{GUTIERREZZABALLA2024103242} is filled by the bit-flip, increasing its value above unity.
For example, in the Sig-based model, 75\% of layer 2 parameters have a `0' in bit 26 and 37.5\% of those are one bit-flip away from having the partial exponent filled.
Similarly, in the HSig-based DNN, 46\% of layer 5 parameters have a `0' in bit 26 and 37.28\% of those are one bit-flip away from having the partial exponent filled.
The corresponding error peaks can be seen in Figures \ref{fig:bitFlipErrorSigmoidNotPruned} and \ref{fig:bitFlipErrorHardSigmoidNotPruned}.

As for the architecture-dependent situations, the scale (gamma) parameter of Batch Normalization layers (dark blue) is the most sensitive factor because its values are usually near $1$.
Besides, due to the skip connections of the U-Net, the first layers of the encoder and the last layers of the decoder are the most sensitive ones.

There are notable AF-specific characteristics.
The mean and standard deviation error rates per parameter and per bit differ significantly: ReLU-based (4.21\%/16.91\%), Sig-based (2.55\%/10.54\%), and HSig-based (2.30\%/9.57\%) DNNs.
High standard deviations stem from error distribution across different DNN positions, with a few locations experiencing severe errors while most remain low, as shown in Figures \ref{fig:bitFlipErrorReLUNotPruned}, \ref{fig:bitFlipErrorSigmoidNotPruned}, and \ref{fig:bitFlipErrorHardSigmoidNotPruned}.
Squashed AFs help contain activation values, preventing unbounded error propagation.
It can also be observed that Sig and HSig-based DNNs have fewer critical parameters compared to the ReLU-based AF.
This suggests that combining bounded AFs with selective modular redundancy or duplication could be a promising approach, potentially improving robustness while balancing resource limitations.
As mentioned in \cite{10769119}, almost all errors produced by bit-flips in the MSB of the exponent occur in parameters with original values in the $1 < |x| < 2$ range, which turn into a $NaN$ (Figures \ref{fig:bitFlipErrorSigmoidNotPruned} and \ref{fig:bitFlipErrorHardSigmoidNotPruned}).
The difference between the mean error rate associated with a bit-flip in bit 30 and the proportion of parameters in the $1 < |x| < 2$ range is much smaller in the squashed AF-based DNNs (17.85\% and 14.48\% in the Sig-based and 15.02\% and 11.42\% in the HSig-based) than in the ReLU-based DNN (35.45\% and 8.12\%).

\begin{figure}[b]
\centerline{\includegraphics[height =6.3cm]{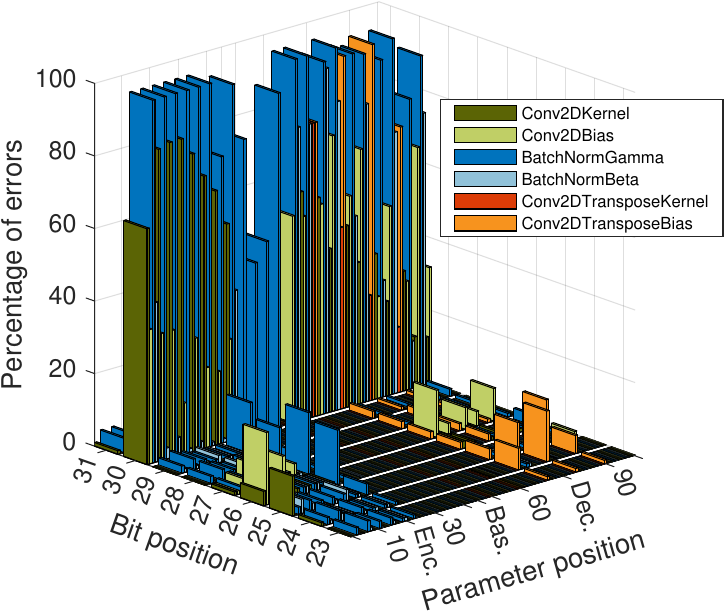}}
\caption{Bit-flip error rate for the pruned ReLU-based DNN.}
\label{fig:bitFlipErrorReLUPruned}
\end{figure}

Errors from a bit-flip in bit 30 in the biases of the final convolution layers are estimated based on their sign.
The bias values are: ReLU-based $(-0.85, 0.32, -0.03, 0.04, -0.17, 0.11)$, Sig-based $(-7.83, 0.32, 0.09, 0.10, 0.04, 0.10)$ and HSig-based $(-7.24, 0.28, -0.00, 0.51, 0.02, 0.06)$.
A flip in a negative bias makes the corresponding class to never be predicted, causing high error rates if the class frequency is high in an image, while a flip in a positive bias forces the class to always be predicted.
Taking that into account, the expected error rates (see Table \ref{tab:expectedErrorRatesNotPruned}) are calculated multiplying the probabilities of the faultless model predicting each class and the probability of a flip occurring in bias $j$, assumed to be $\frac{1}{6}$ (error model from \cite{GUTIERREZZABALLA2024103242}).
These values are confirmed experimentally (34\%, 70\% and 53\%) as depicted in Figures \ref{fig:bitFlipErrorReLUNotPruned}, \ref{fig:bitFlipErrorSigmoidNotPruned} and \ref{fig:bitFlipErrorHardSigmoidNotPruned}.
The slight rate differences are a result of the FI not resulting in the same number of bit-flips for each class.

\begin{table}[h!]
\centering
\resizebox{8cm}{!}{
\begin{tabular}{c|c|c|c|c|c|c|c|c|}
\cline{2-9}
 & \textbf{$P^j_{fi}$} & \textbf{$P^0_{m}$} & \textbf{$P^1_{m}$} & \textbf{$P^2_{m}$} & \textbf{$P^3_{m}$} & \textbf{$P^4_{m}$} & \textbf{$P^5_{m}$} & \textbf{$err_{30}$} \\ \hline
\multicolumn{1}{|c|}{\textbf{ReLU}}  & 1/6 & 0 & 55.09 &  4.41 & 73.05 &  7.47 & 83.73 & 37.29\% \\ \hline
\multicolumn{1}{|c|}{\textbf{Sig.}}  & 1/6 & 0 & 56.80 & 95.11 & 75.63 & 92.17 & 80.28 & 66.66\% \\ \hline
\multicolumn{1}{|c|}{\textbf{HSig.}} & 1/6 & 0 & 58.66 &  4.71 & 75.72 & 93.63 & 76.71 & 51.57\% \\ \hline
\end{tabular}}
\caption{Expected error in noncompressed DNNs according to error model from \cite{GUTIERREZZABALLA2024103242}.
$P^j_{fi}$ is is the prob. of a flip being in bias j, 1/6, and $P^j_{m}$ is the prob. of the faultless model predicting $m$ as the class.}
\label{tab:expectedErrorRatesNotPruned}
\end{table}

\begin{figure}[t]
\centerline{\includegraphics[height =6.3cm]{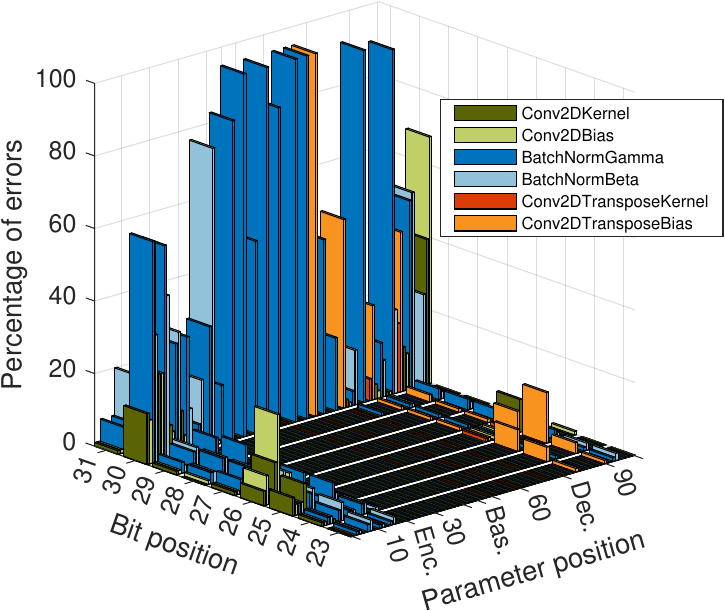}}
\caption{Bit-flip error rate for the pruned Sig.-based DNN.}
\label{fig:bitFlipErrorSigmoidPruned}
\end{figure}

\begin{figure}[b]
\centerline{\includegraphics[height =6.3cm]{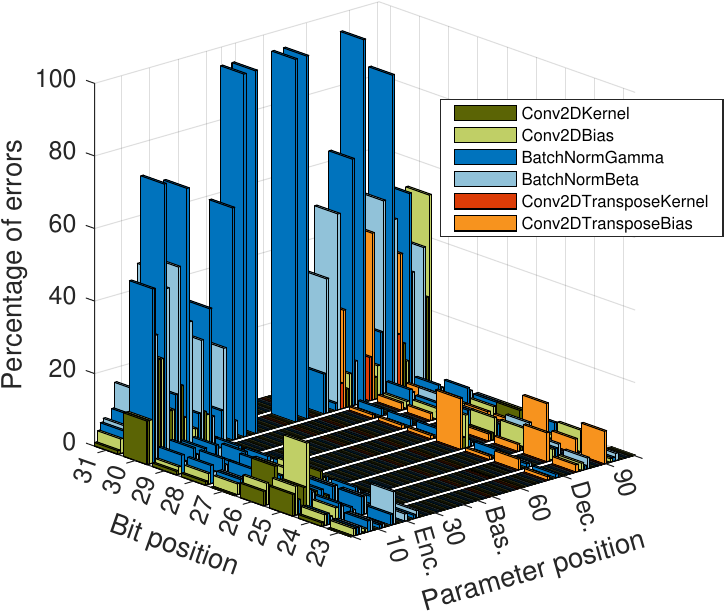}}
\caption{Bit-flip error rate for the pruned HSig.-based DNN.}
\label{fig:bitFlipErrorHardSigmoidPruned}
\end{figure}

\vspace{-0.8cm}
\subsection{Pruned models}
DNNs leverage their overparameterized and parallel structure to represent features redundantly across multiple filters and layers.
However, aggressive pruning reduces this redundancy, leading to a more compact yet potentially less expressive model.
With fewer parameters, the ability to capture complex patterns may be constrained.
Nevertheless, the pruning process was limited to a maximum loss of 1.5 points in GIoU and WIoU, ensuring essential feature representation was preserved.
The evaluation of the three pruned, nonquantized models on the test set has produced the following metrics: (94.46/88.48) for the ReLU, (92.45/83.36) for the Sig and (92.43/82.80) for the HSig.

A quantitative analysis comparing bit-flip error rates before and after pruning (Figure \ref{fig:bitFlipErrorReLUNotPruned} with \ref{fig:bitFlipErrorReLUPruned}, Figure \ref{fig:bitFlipErrorSigmoidNotPruned} with \ref{fig:bitFlipErrorSigmoidPruned}, and Figure \ref{fig:bitFlipErrorHardSigmoidNotPruned} with \ref{fig:bitFlipErrorHardSigmoidPruned}) shows that while all mean and standard deviation error rates increased, the most pruned DNNs experienced the highest growth: the ReLU-based DNN has suffered an increase from 4.21\% to 6.16\% and from 16.91\% to 19.91\%, the HSig-based DNN from 2.30\% to 3.36\% and from 9.57\% to 11.39\%, but the Sig-based DNN, the least pruned DNN, has increased only from 2.55\% to 2.87\% and from 10.54\% to 12.00\%.
More overparameterized models are less affected by soft errors, as faults are more likely to impact redundant parameters.
However, pruning reduces these redundancies, making errors more impactful.
On the other hand, the reduced memory footprint of pruned models make them statistically less sensitive to SEUs.
This trade-off between compression and robustness must be considered when evaluating DNN fault tolerance.
The difference between the mean error rate associated with a bit-flip in bit 30 and the proportion of parameters in the $1 < |x| < 2$ range further highlights this effect: 50.95\% and 12.69\% for the ReLU-based DNN, 20.58\% and 16.58\% for the Sig-based DNN and 20.32\% and 14.79\% with the HSig-based DNN.

Using the same approach, since finetuning did not alter bias values (rounded to two decimals), Table \ref{tab:expectedErrorRatesPruned} presents the error rates caused by a flip in bit 30 of the final convolution layer's biases.
These values mirror the 32\%, 68\% and 52\% error rates from Figures \ref{fig:bitFlipErrorReLUPruned}, \ref{fig:bitFlipErrorSigmoidPruned} and \ref{fig:bitFlipErrorHardSigmoidPruned}.
The overall error rate trend remains unchanged, with the most significant errors caused by bit-flips in the MSB of the exponent.
Additionally, secondary error peaks appear in the remaining exponent and sign bits outside the base of the U-Net.
This suggests that training larger, more sparse models could enhance DNN resilience against SBUs.
This practice, however, would hinder DNN performance when deployed on edge devices.

\begin{table}[h!]
\centering
\resizebox{8cm}{!}{
\begin{tabular}{c|c|c|c|c|c|c|c|c|}
\cline{2-9}
 & \textbf{$P^j_{fi}$} & \textbf{$P^0_{m}$} & \textbf{$P^1_{m}$} & \textbf{$P^2_{m}$} & \textbf{$P^3_{m}$} & \textbf{$P^4_{m}$} & \textbf{$P^5_{m}$} & \textbf{$err_{30}$} \\ \hline
\multicolumn{1}{|c|}{\textbf{ReLU}}  & 1/6 & 0 & 55.66 &  4.35 & 72.93 &  7.37 & 83.13 & 37.24\% \\ \hline
\multicolumn{1}{|c|}{\textbf{Sig.}}  & 1/6 & 0 & 58.14 & 95.30 & 76.52 & 93.30 & 76.75 & 66.67\% \\ \hline
\multicolumn{1}{|c|}{\textbf{HSig.}} & 1/6 & 0 & 57.80 &  5.01 & 76.05 & 93.66 & 77.51 & 51.67 \% \\ \hline
\end{tabular}}
\caption{Expected error in pruned DNNs according to error model from \cite{GUTIERREZZABALLA2024103242}.
$P^j_{fi}$ is is the prob. of a flip being in bias j, 1/6, and $P^j_{m}$ is the prob. of the faultless model predicting $m$ as the class.}
\label{tab:expectedErrorRatesPruned}
\end{table}

\begin{figure}[b]
\centerline{\includegraphics[height =6.4cm]{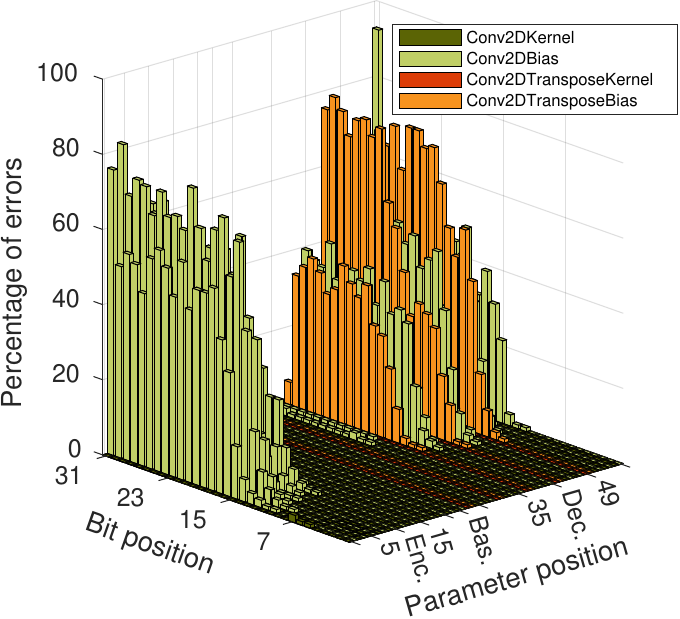}}
\caption{Bit-flip error rate for the pruned quantized ReLU-based DNN.}
\label{fig:bitFlipErrorPrunedQuantizedReLU}
\end{figure}

\vspace{-0.8cm}
\subsection{Pruned quantized models}
\begin{figure}[t]
\centerline{\includegraphics[height =6.3cm]{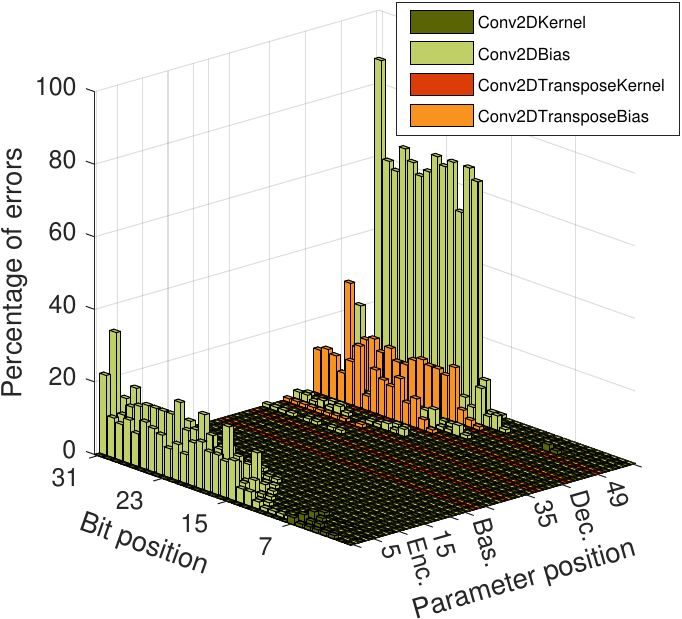}}
\caption{Bit-flip error rate for the pruned quantized Sig-based DNN.}
\label{fig:bitFlipErrorPrunedQuantizedSigmoid}
\end{figure}

\begin{figure}[b]
\centerline{\includegraphics[height =6.3cm]{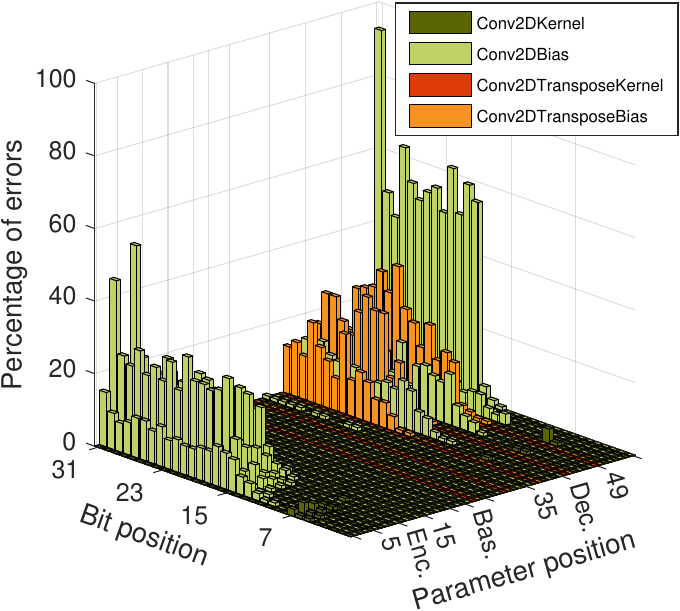}}
\caption{Bit-flip error rate for the pruned quantized HSig-based DNN.}
\label{fig:bitFlipErrorPrunedQuantizedHardSigmoid}
\end{figure}

In the last experiment, we evaluated the combined effects of pruning and integer quantization.
The error rates for the ReLU-based, Sig-based and HSig-based DNNs are shown in Figures \ref{fig:bitFlipErrorPrunedQuantizedReLU}, \ref{fig:bitFlipErrorPrunedQuantizedSigmoid} and \ref{fig:bitFlipErrorPrunedQuantizedHardSigmoid}.
While they may appear less robust than nonquantized DNNs, only biases remain as sensitive parameters and, since they constitute a small proportion of the DNN parameter set, the actual amount of sensitive parameters is much lower.
Bounded AFs consistently enhance robustness with the Sig-based DNN (0.14\% and 7.79\%) and the HSig-based DNNs (0.24\% and 10.93\%) showing lower mean and mean nonzero error rates than ReLU-based DNN (0.76\% and 28.23\%).

The error rate in the bias of the final convolution layer can also be estimated but, as a consequence of the integer representation, the approach differs.
Once a bit-flip has produced a sufficient increase to cause an output error, additional changes in higher bit positions have no further impact ([17-30] for ReLU-based and [19-30] for Sig-based and HSig-based DNNs).
To estimate a single statistical error rate from upsets across all bit positions, a linear weighting approximation based on bit significance has been applied as done in \cite{GUTIERREZZABALLA2024103242}.
The values, 37.89\% for ReLU-based, 61.52\% for Sig-based and 52.49\% for HSig-based DNNs, are very similar to the expected error rates (see Table \ref{tab:expectedErrorRatesPrunedQuantized}).

\begin{table}[h!]
\centering
\resizebox{8cm}{!}{
\begin{tabular}{c|c|c|c|c|c|c|c|c|}
\cline{2-9}
 & \textbf{$P^j_{fi}$} & \textbf{$P^0_{m}$} & \textbf{$P^1_{m}$} & \textbf{$P^2_{m}$} & \textbf{$P^3_{m}$} & \textbf{$P^4_{m}$} & \textbf{$P^5_{m}$} & \textbf{err.} \\ \hline
\multicolumn{1}{|c|}{\textbf{ReLU}}  & 1/6 & 0 & 55.68 &  4.24 & 73.03 &  6.96 & 82.48 & 37.06 \% \\ \hline
\multicolumn{1}{|c|}{\textbf{Sig.}}  & 1/6 & 0 & 59.09 & 95.48 & 77.43 & 94.07 & 73.93 & 66.66 \% \\ \hline
\multicolumn{1}{|c|}{\textbf{HSig.}} & 1/6 & 0 & 58.78 &  5.23 & 76.03 & 93.82 & 76.61 & 51.75 \% \\ \hline
\end{tabular}}
\caption{Expected error in compressed DNNs according to error model from \cite{GUTIERREZZABALLA2024103242}.
$P^j_{fi}$ is is the prob. of a flip being in bias j, 1/6, and $P^j_{m}$ is the prob. of the faultless model predicting $m$ as the class.}
\label{tab:expectedErrorRatesPrunedQuantized}
\end{table}

\section{Deployment characterization and discussion}\label{sec:characterization}
\begin{figure}[t]
\centerline{\includegraphics[height = 4cm]{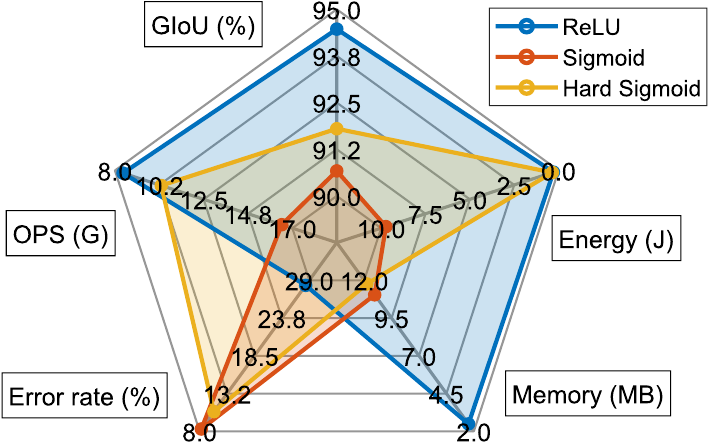}}
\caption{Deployment performance comparison between ReLU-based (blue), Sig-based (red) and HSig-based (yellow) DNNs.}
\label{fig:FoM}
\end{figure}

All designed DNNs were implemented on an AMD-Xilinx K26 SoM, which features an XCK26-SK-KV260-G Zynq UltraScale+ MPSoC containing a 64-bit Quad-Core ARM A-53 processor and a 16nm FinFET Programmable Logic FPGA.
The DNNs were quantized to 8-bits and deployed on a single-core B4096 Deep Processing Unit (DPU) using AMD-Xilinx's Vitis AI 3.5 environment.
Figure \ref{fig:FoM} shows the comparison in terms of segmentation metrics (IoU), memory footprint (MB), computational complexity (GOPS), error resilience (error rate) and efficiency (J).

For GIoU/WIoU, performance remains close to noncompressed models, with ReLU-based (94.47/88.37) outperforming Sig-based (90.67/80.96) and HSig-based (91.80/80.72) models.
ReLU’s arithmetic simplicity and higher compressibility lead to the highest throughput (22.68 FPS), followed by the HSig-based model (19.83 FPS).
Since Sig AF cannot be computed on the DPU, its performance drops significantly (0.59 FPS), though it achieves the lowest mean power consumption (5.74 W) compared to ReLU-based (6.62 W) and HSig-based (6.90 W) models.
In terms of energy efficiency, the ReLU-based and HSig-based models are the most efficient (0.29 J and 0.35 J, respectively) while, in this deployment, Sig-based model is the most energy-consuming (9.73 J).
AF choice depends heavily on the application’s constraints.
If error resilience is prioritized, squashed AFs (Sig-based and HSig-based models with error rates of 7.79\% and 10.93\%) are preferable over ReLU (28.23\%), which is less robust.
Those requirements shape the vertices of Figure \ref{fig:FoM} but, in this general scheme, axes limits were set to the nearest integers below the minimum and above the maximum measured values.

Although not the main focus of this article, it is important to explain how pruning translates to lower FPGA hardware costs.
The benefits are twofold.
First, reducing the number of operations and parameters speeds up DNN inference, improving throughput and lowering latency.
Second, considering the specific characteristics of a general pipelined application where the image preprocessing stage is the bottleneck, the same throughput can be obtained by utilizing a smaller DPU configuration, which reduces FPGA resource usage.

\section{Conclusions}\label{sec:conclusions}
This article explores the use of bounded AFs in image segmentation DNNs to evaluate their robustness against soft errors on embedded platforms for safety-critical applications.
In 32-bit floating-point models, a critical situation arises when the MSB of the exponent flips for values in the range $1 < |x| < 2$, causing a conversion to $NaN$, which mainly impacts gamma parameters in Batch Normalization layers.
Aside from that situation, which is independent of the selected AF, bounded functions improve model resilience against single bit-flips, since they limit the propagation of error perturbations across layers.
Since the models based on bounded AFs have fewer critical parameters compared, this approach could be combined with using selective memory hungry methods such as modular redundancy, but just focusing in a considerable smaller number of parameters.

Pruned models experience a loss in robustness inherent of overparameterized nature.
Nevertheless, smaller models generally require fewer logic and memory resources, reducing the probability of soft errors.
Therefore, the decision to apply pruning techniques must be weighed against their impact on robustness and the specific processor design for deployment.
In addition to this, ReLU-based DNN has shown the greatest pruneability achieving a reduction of 99\% in the number of parameters and of 75\% in the number of OPS, so there is a trade-off between the resilience offered by bounded AFs and the need for aggressive model compression to enhance computational performance and minimize memory usage.
For fully pruned and quantized integer models, the binary representation prevents the appearance of $NaNs$ and infinities, substantially lowering error rates while preserving the same relative robustness patterns among AFs.

When deployed on the AMD-Xilinx KV260 SoM, the ReLU-based DNN achieved the best IoU and demonstrated the highest efficiency in throughput and power consumption.
However, the use of HSig AFs emerges as a viable alternative, offering a good balance of robustness and performance.
Even Sig AFs could be considered to get the best error resilience if accelerated by hardware.
In conclusion, while bounded AFs enhance robustness, careful evaluation of the trade-off between resilience, efficiency, and model compression is essential for deployment in safety-critical embedded applications.

\begin{acks}
This work has been partially supported by the UPV/EHU under grant GIU21/007 and by the Basque Governments under grants PRE\_2024\_2\_0154 and KK-2023/00090.
\end{acks}

\balance

\bibliography{iet-ell}
\bibliographystyle{iet}

\end{document}